\documentclass[letterpaper]{article} % DO NOT CHANGE THIS
\usepackage{aaai25}  % DO NOT CHANGE THIS
\usepackage{times}  % DO NOT CHANGE THIS
\usepackage{helvet}  % DO NOT CHANGE THIS
\usepackage{courier}  % DO NOT CHANGE THIS
\usepackage[hyphens]{url}  % DO NOT CHANGE THIS
\usepackage{graphicx} % DO NOT CHANGE THIS
\usepackage{natbib}  % DO NOT CHANGE THIS AND DO NOT ADD ANY OPTIONS TO IT
\usepackage{caption} % DO NOT CHANGE THIS AND DO NOT ADD ANY OPTIONS TO IT
\usepackage{microtype}

\usepackage{graphicx}
\usepackage{float}

\usepackage[ruled,vlined]{algorithm2e}
\usepackage{algorithmic}
\algsetup{linenosize=\small}

\usepackage{enumitem}
\setlist[itemize]{leftmargin=*, topsep=0pt, noitemsep}

\usepackage{amssymb}
\usepackage{amsmath}

\usepackage{booktabs,tabularx}

\usepackage{soul}

\usepackage{xcolor, colortbl}
\newcommand{\hlc}[2][yellow]{{%
    \colorlet{foo}{#1}%
    \sethlcolor{foo}\hl{#2}}%
}

\usepackage{newfloat}
\usepackage{listings}
\DeclareCaptionStyle{ruled}{labelfont=normalfont,labelsep=colon,strut=off} % DO NOT CHANGE THIS
\floatstyle{ruled}
\newfloat{listing}{tb}{lst}{}
\floatname{listing}{Listing}
\title{Baby Bear: Seeking a Just Right Rating Scale for Scalar Annotations}
\author{
    Xu Han\textsuperscript{\rm 1}\thanks{Work done at Johns Hopkins University},
    Felix Yu\textsuperscript{\rm 2},
    João Sedoc\textsuperscript{\rm 3},
    Benjamin Van Durme\textsuperscript{\rm 2}
}
\affiliations{
    \textsuperscript{\rm 1}Yale University,
    \textsuperscript{\rm 2}Johns Hopkins University,
    \textsuperscript{\rm 3}New York University\\
    \textsuperscript{\rm 1}xu.han.xh365@yale.edu,
    \textsuperscript{\rm 2}\{fyu17,vandurme\}@jhu.edu,
    \textsuperscript{\rm 3}jsedoc@stern.nyu.edu
}

\begin{document}

\maketitle

\begin{abstract}
Our goal is to identify a mechanism for efficiently assigning scalar ratings to each of a large set of elements. For example, \textit{``what percent positive or negative is this product review?''}  When sample sizes are small, prior work has advocated for methods such as Best Worst Scaling (BWS) as being more robust than direct ordinal annotation ("Likert scales"). Here we first introduce  IBWS, which \textbf{\ul{i}}teratively collects annotations through \textbf{\ul{B}}est-\textbf{\ul{W}}orst \textbf{\ul{S}}caling, resulting in robustly ranked crowd-sourced data. While effective, IBWS is too expensive for large-scale tasks. Using the results of IBWS as a best-desired outcome, we evaluate various direct assessment methods to determine \emph{which are both cost-efficient and best correlates to a large scale BWS annotation strategy}.
Finally, we illustrate in the domains of dialogue and sentiment analysis how these annotations can drive robust learning-to-rank models for automated assessment.

% Conditioned on some query, such as \textit{``how positive or negative is this review?''}, our goal is a mechanism for efficiently assigning scalar values to each of a large set of elements. Given the annotator inconsistency and varying response time, choosing an appropriate scale (e.g., Likert or slider) to elicit reliable scalar annotations is often challenging when designing surveys. To address this, we leverage the robust nature of Best-Worst Scaling protocol and introduce a ranking algorithm called IBWS that \textbf{\ul{i}}teratively collects annotations through \textbf{\ul{B}}est-\textbf{\ul{W}}orst \textbf{\ul{S}}caling, resulting in robust ranked crowd-sourced data. While effective, IBWS is too expensive for large-scale tasks. To find a more practical alternative, we compare different $\mathcal{O}(N)$ direct assessment methods and identify simple slider protocol as the most reliable, and efficient. Empirically, we find it closely aligns with the IBWS rankings and the ground truth than more intensive approaches. 
% We further train learning-to-rank models to predict the annotations automatically, demonstrating their effectiveness on two tasks: sentiment analysis and rating dialogue interactions.

% We introduce a vertical-drag protocol and compare the LTR models trained on the best scalar interface to the proposed interface. 
% Our findings indicate that the slider interfaces are not only most closely correlated with both the IBWS rankings and the original labels on a product review sentiment task but also have the fastest response time.
\end{abstract}


\section{Introduction}
\label{sec:intro}
% The task of ranking items is an integral part of many machine learning problems, such as search engine result, computer vision, machine translation n-best list, and language generation system rankingaa
% such as ordering search engine results, assessing computer vision outputs, selecting the top translation in machine translation, and evaluating language generation systems 
Human annotations are crucial for improving model performance. With the rise of large language models (LLMs), the demand for large-scale human annotations has grown, particularly for pre-training, supervised fine-tuning (SFT) and incorporating human feedback in the rewards function (RLHF)~\citep{devlin2019bertpretrainingdeepbidirectional,chen2024selfplayfinetuningconvertsweak,liang2024richhumanfeedbacktexttoimage}. However, gathering reliable human annotations at scale is both expensive and time-consuming, making it crucial to develop strategies that can reduce these costs while ensuring the data's reliability. Additionally, many machine learning tasks—such as web search, computer vision, recommender systems, dialogue systems, and machine translation—rely on models that can effectively rank items or responses \cite{liu2009learning,weston2010large}. Learning-to-rank (LTR) models, in particular, require training data with accurate rankings of large item sets,  which can be challenging to obtain. To address this challenge, recent progress generally falls along two lines: \emph{optimizing annotation protocols or improving LTR models}. 

% However, gathering reliable human annotations at scale is both expensive and time-consuming, making it crucial to develop strategies that can reduce these costs while ensuring the data's reliability. 
% Despite its importance, there has been limited research aimed at discovering such annotation protocols. 
Under the first taxonomy, many efforts have been made to develop more effective annotation protocols that either produce higher-quality annotations or minimize the number of human annotations required~\cite{efficient-online-scalar-annotation-with-bounded-support,mohankumar2022activeevaluationefficientnlg, mishra2022crosstaskgeneralizationnaturallanguage, lee-etal-2023-common}. 
% NLP researchers aim to optimize annotation protocols, annotation sample efficiency, and instructions ~\cite{mohankumar-khapra-2022-active, sakaguchi-van-durme-2018-efficient, mishra-etal-2022-cross, lee-etal-2023-common}. 
However, this paradigm often fails to consider the connection between the collected annotations and subsequent learning-to-rank processes. For instance, Best Worst Scaling (BWS) can generate relative rankings within a small set of items: humans incrementally pick the best and worst in a small set, under some category. Yet BWS is expensive if seeking a global ranking across a large set of items. 
On the other hand, model-in-the-loop ranking focuses on enhancing the model's ranking ability by redesigning its structure but can overlook the quality of the annotations~\cite{xia2008listwise, liu2009learning,shah2016simplerobustoptimalranking}. Our goal is to bridge this gap by identifying an annotation protocol that not only efficiently produces robust ranked annotations but can also be used to train an LTR model to predict rankings.
% that offers both the highest correlation between the learned model and ground truth, as well as optimal annotation efficiency.

Motivated by the fact that BWS is more effective than direct ordinal annotations~\citep{louviere_flynn_marley_2015}, in this study, we first introduce IBWS (Iterated Best-Worst Scaling), a novel ranking algorithm designed to generate reliable annotations by iteratively refining feedback from BWS. Although we show that IBWS is effective, its complexity makes it challenging for large-scale tasks. To address this, we evaluate various direct assessment methods and find that a simple slider protocol as the most reliable and efficient alternative using the results of IBWS as a best-desired outcome. Empirically, we demonstrate that a slider protocol closely aligns with IBWS rankings and ground truth. Furthermore, we train  LTR models with collected slider annotations to automatically predict rankings, which is tested on two tasks: \emph{sentiment analysis and rating dialogue interactions}. Our results highlight the effectiveness of the LTR models, which not only enhances the accuracy of model predictions but also reduces the time and cost associated with data collection, offering a scalable solution for many ML applications.

The main contributions of this study are:
\begin{itemize}
    \item We propose IBWS, an effective annotation collection algorithm that generates robust ranked annotations. To facilitate BWS annotations and empirically analyze the effectiveness of IBWS, we develop two interfaces: a standard two-column BWS interface and a vertical-drag interface; 
    \item To find a more practical alternative, we compare different $\mathcal{O}(N)$ direct assessment methods and identify the simple slider protocol as the most reliable, and efficient; 
    % Empirically, we find it closely aligns with the IBWS rankings and the ground truth than more intensive approaches. 
    \item We further train LTR models to predict the annotations automatically, demonstrating their effectiveness on two tasks: sentiment analysis and rating dialogue interactions. 
\end{itemize}

\begin{figure}[t]
\centering
\includegraphics[width=1\linewidth]{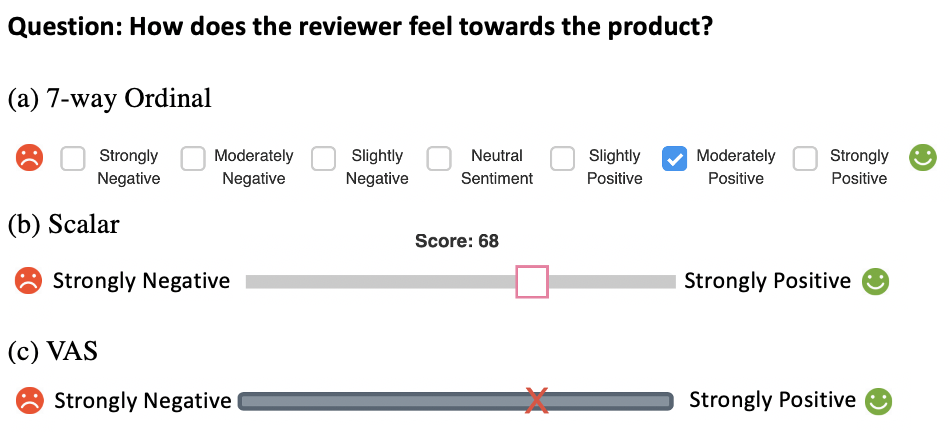}
\caption{Direct assessment protocols for sentiment.}
\label{fig:single}
\end{figure}

\section{Background}

Three approaches are frequently used in surveys for sentiment data collection: direct assessment, pairwise ranking, and best-worst scaling (BWS). 

\paragraph{Direct Assessment} of scalar annotation (Figure \ref{fig:single}) is widely favored for its simplicity and ease of analysis. One of the most favored protocols is the \textbf{n-way ordinal scale} (a.k.a., Likert scale)~\citep{likert_1987} in which annotators select from a range of ordered labels. However, discrete scales can lead to inaccurate judgments when an annotator's opinion falls between two points on the scale \citep{belz-kow-2011-discrete}. Additionally, the sentiment range each label represents can be unclear; for example, in a 7-point ordinal scale (shown in Figure \ref{fig:single}(a)), the distance between \emph{moderately negative} and \emph{strongly negative} may not correspond to the distance between \emph{moderately negative} and \emph{slightly negative}.

This issue can be mitigated by using \emph{continuous scales} like the \textbf{Slider} and \textbf{Visual Analog Scale (VAS)}.  Instead of choosing from discrete labels, annotators can indicate a precise value on a scale, typically ranging from 0 to 100. This protocol may introduce bias, as it requires an initial location of the slider on the scale which is then adjusted by the annotator \citep{RePEc:taf:mpopst:v:25:y:2018:i:2:p:112-122}. To counteract this the VAS protocol provides a blank line, requiring the annotator to select ("click on") a location.
%It was reported in literature that sliders requires more response time than VAS \citep{webexperiment}. 

A common challenge in direct assessment is \emph{the lack of calibration between annotators}. For instance, reviewer A might give a 5-star rating to any product they find \emph{not bad}, while reviewer B reserves a 5-star rating only for products that exceed expectations~\citep{jansen1984ridit}. Also, direct assessment suffer from \emph{high variance} and \emph{sequence effects}~\citep{mohankumar2022activeevaluationefficientnlg}, highlighting the need for a more robust data collection interface.

\begin{figure}[!t]
\centering
\includegraphics[width=1\linewidth]{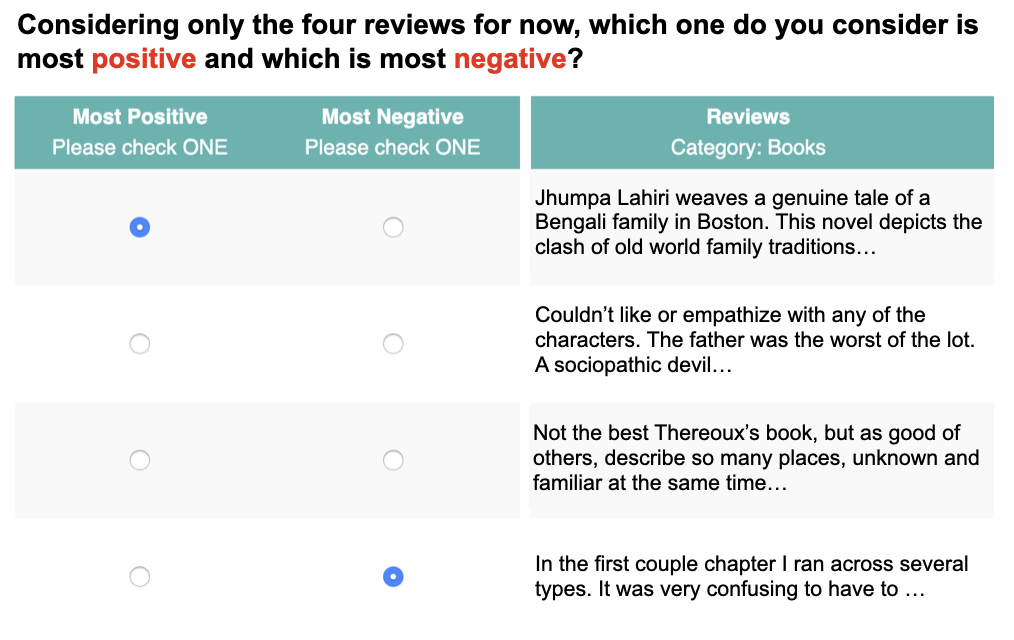}
\caption{BWS protocol on Amazon review sentiment.}
% \vspace{-2mm}
\label{fig:bws}
\end{figure}

\begin{figure}[t]
\centering
\includegraphics[width=1\linewidth]{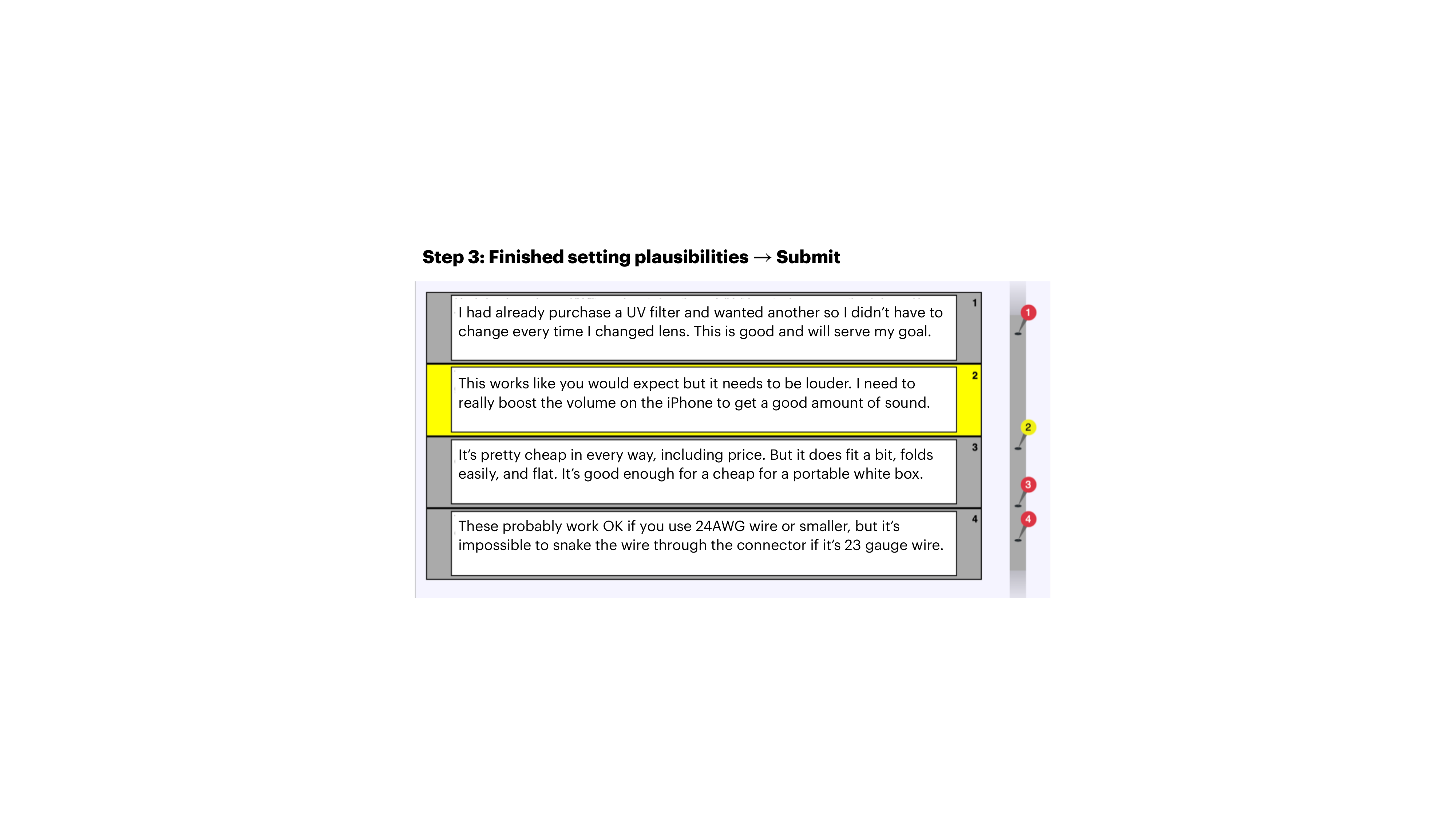}
\caption{Vert-drag BWS interface.}
% \vspace{-4mm}
\label{fig:IBWS-VD}
\end{figure} 

\paragraph{Pairwise Ranking} 
% asks annotators to rank items in a list. However, this approach becomes impractical for larger lists. BWS offers an alternative by having annotators identify only the best and worst items from a list, which are then used to rank the entire set. 
compares two items at a time to determine which one is more positive or negative, making it simpler and less cognitively demanding for annotators. Although this reduces individual judgment errors, it can be time-consuming and inefficient for large datasets due to the need for $O(n^2)$ comparisons.

\paragraph{Best-Worst Scaling (BWS)} also known as MaxDiff sorting~\citep{louviere_flynn_marley_2015}, presents annotators with sets of $n$-tuple items and asks them to select the best and worst items in each set. Typically, BWS is conducted with $n = 4$ items per set (Figure \ref{fig:bws}), as recommended by \citet{kiritchenko-mohammad-2017-best}. Instead of categorizing ordinal labels or assigning scores, BWS allows annotators to directly compare items, which simplifies the task and reduces inter-annotator variability, leading to more consistent annotations. However, BWS is resource-intensive and time-consuming, requiring more human interactions and taking approximately 4.5 times longer than categorical annotation \citep{glenn-etal-2022-viability}.
% Consequently, our work seeks to develop a direct scalar protocol that minimizes the number of interactions needed to approximate a BWS-based ranking.

\section{Methods}
\label{Methodology}

We begin this section by introducing IBWS algorithm. Then, we discuss an LTR model to predict annotations. 
 
% which goes first
% IS THIS REPEATED WITH IBWS METHOD PART?
% Our IBWS experiment is illustrated by Figure \ref{fig:bws_explanation}. To set up the experiment, we first randomly select four instances from 1k reviews per product type. We manually annotate the most positive and most negative reviews of these four as \emph{max} and \emph{min}, respectively. The \emph{max} and \emph{min} reviews are repeatedly grouped with two more random reviews to form new 4-tuple instances. Then we ask annotators to use BWS for each 4-tuple instance. After an iteration, by using the IBWS algorithm, reviews are eventually split into three buckets. For each produced bucket, we rerun the experiment and produce another three buckets. 

% To sort reviews sufficiently fine-grained, we set the number of iterations to 3 with the reviews partitioned into 27 buckets. It is important to note that, for each 4-tuple instance, when the worker-annotated most positive $\emph{max}'$ and negative $\emph{min}'$ reviews are different from the anchored \emph{max} and \emph{min} (manually annotated by researcher), the worker will be dynamically asked to compare \emph{max} or \emph{min} with the non-$\emph{max}'$ / $\emph{min}'$ elements (in two-column IBWS interface). An example is presented in Appendix \ref{sec:appendix}. 

\subsection{Iterated Best-Worst Scaling}
\begin{figure}[t]
\begin{center}
\includegraphics[width=1\linewidth]{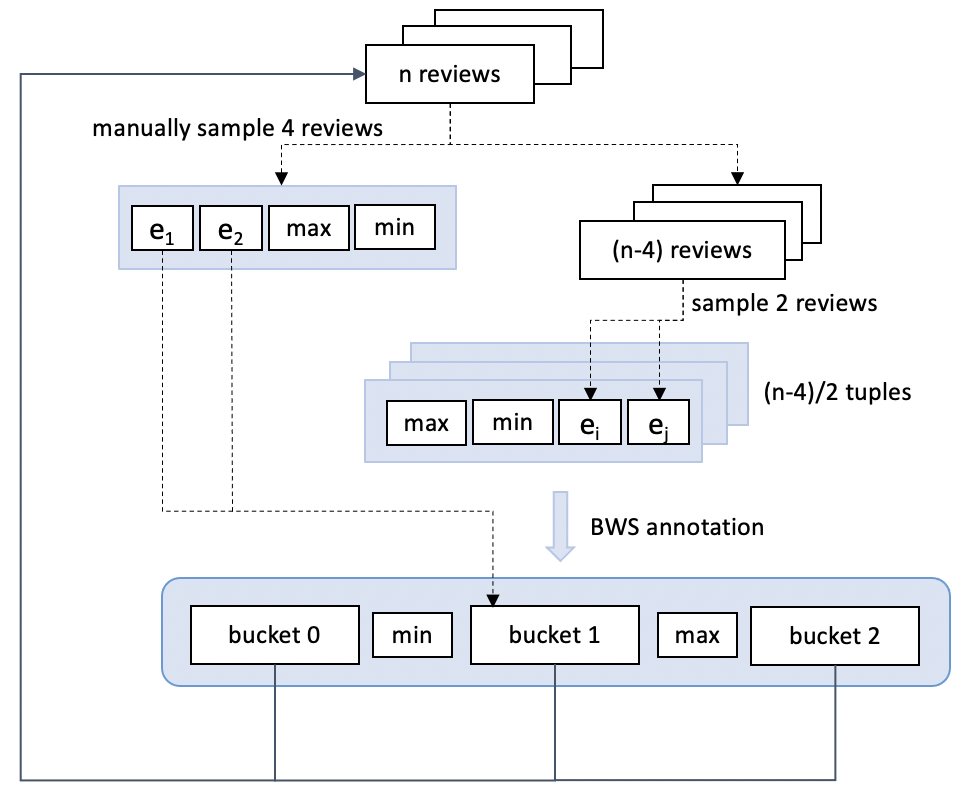}
\end{center}
\caption{An illustration of IBWS algorithm.}
\label{fig:bws_explanation}
\end{figure}

To perform crowd-sourced ranking on BWS annotations, we develop the IBWS algorithm as explained in Algorithm \ref{alg:IBWS}. Inspired by Quicksort \citep{Hoare1961Algorithm6P}, we implement ranking by iteratively collecting annotations using BWS. We first assign all items to a single bucket from which we randomly sample 4 items without replacement. We manually label the best (\emph{max}) and worst  (\emph{min}) elements. Motivated by quicksort comparisons on a single pivot, we perform BWS for every remaining element in the bucket: annotators are repeatedly given 4-tuples consisting of \emph{max}, \emph{min} and two randomly selected items, to then select a new \emph{max'}, \emph{min'}. This allows us to rank the two new items relative to the initial pair. The algorithm results in a multiplicative 3-way partition of the data after each iteration (\emph{buckets 0,1,2} as shown in Figure \ref{fig:bws_explanation}). After  $k$ iterations all items are placed in one of $3^k$ buckets. This can be considered a fine-grain ordinal scale. For example, 4 iterations of this approach leads to each element being assigned to one of 81 ordered buckets (ordinal labels).

%\paragraph{Time Complexity} 
%% Empirically based on observations during development, we believe 
%Given N items, IBWS finishes ranking in $\mathcal{O}(N \log N)$. At each iteration, we annotate the entire dataset with BWS in $\mathcal{O}(N)$ running time. There's a minimum granularity that humans can differentiate no matter how many items there are to be sorted, so in practice there's a constant number of iterations needed to reach this granularity. To fully sort N items, the number of iterations needed is $k = \log N $, given the same logic as quicksort.  
% If fixing the number of iteration, the complexity will go down to $\mathcal{O}(N)$. However, considering the size of large datasets, the costs of increasing k can be expensive.
% Future work could explore this further\footnote{To make the algorithm more precise, we can look at when annotators start to disagree in their judgements at some iteration. That is, the "terminal iteration" would be when annotators are regularly making different judgements or saying "can't tell the difference between these elements, they are the same under the given scale".}. 
% For example, when N = 100k, ranking all elements requires k = 6 iterations, indicating a large costs of evaluations.   
% Note that this requires more evaluations than a direct scalar protocol. We address the problem by finding a ``gold'' scalar annotation that best correlates with IBWS ranking but requires fewer evaluations.  

%\noindent
We consider two BWS interfaces to gather annotations:
\vspace{-2mm}
\paragraph{\emph{two-column BWS interface}} A standard BWS interface (Figure~\ref{fig:bws}) that presents four items sided with two columns of buttons for \emph{best} and \emph{worst} respectively \citep{POTOGLOU20111717}. 
% This interface is straightforward to understand but gives little insight into the perceived sentiment distance between reviews. 
\paragraph{\emph{vertical-drag BWS interface}} To better understand how the items are ranked from the annotator's perspective, annotators can indicate the relative sentiment distance between reviews on a vertical bar and rank reviews by dragging them vertically (Figure \ref{fig:IBWS-VD}).

% \paragraph{Researcher Engagement}
% One possible way of decreasing the annotation costs is to incorporate strong researchers' engagement. In particular, when choosing the initial \textit{max} and \textit{min} items, instead of randomly selecting, the researcher can look at a subset of the data and choose these items to roughly split the sets evenly. Similarly to quicksort, a good choice of ``pivots'' reduces the total number of comparisons done.

{\SetAlgoNoLine%
\begin{algorithm}[t]
\algsetup{linenosize=\small}
\scriptsize
    \KwIn{All elements to be partitioned $\{E_1^N\}$}
    \KwOut{ Sorted elements $\{{E'}_{1}^{N}\}$}
    $E'$ = IBWS(E)\;
    \ForEach{${E'}_i \in E'$} {
        IBWS(${E'}_i$)\;
    }
    \SetKwProg{Fn}{Function}{:}{}
    \Fn{IBWS (E)}{
        \eIf{$ |E| < 4 $}
        {$ \mathrm{BWS}(E) $}
        {
            $L, M, U \leftarrow \varnothing $ \;
            $S \leftarrow$ sampling 4 items from $E$ \;
            $s_{\max}, s_{\min}, S_{\rm others} \leftarrow \mathrm{BWS}(S)$ \;
            $M \leftarrow M \cup S_{\rm others}$ , $E' \leftarrow E \setminus S$\;
          \While{$|E'| > 0$}{
                $ e_1, e_2 \leftarrow $ sampling 2 items from $E'$\;
                $ s'_{\max}, s'_{\min} \leftarrow \mathrm{BWS}(\{s_{\max}, s_{\min}, e_1, e_2\})$\;
                \uIf{$s_{\max} \neq s'_{\max}$ {\bf and} $s_{\min} \neq s'_{\min}$}{
                    $U \leftarrow U \cup \{s'_{\max}\}$,
                    $L \leftarrow L \cup \{s'_{\min}\}$\;
                }
                \uElseIf{$s_{\max} \neq s'_{\max}$} {
                    $U \leftarrow U \cup \{s'_{\max}\}$\;
                    $s' \leftarrow S \setminus \{s_{\max}, s'_{\max}, s_{\min}\}$\;
                    \uIf{$s' < s_{\max}$}{$M \leftarrow M \cup \{s'\} $\;}
                    \uElse{$U \leftarrow U \cup \{s'\}$\;} 
                }
                \uElseIf{$s_{\min} \neq s'_{\min}$} {
                    $L \leftarrow L \cup \{s'_{\min}\}$\;
                    $s' \leftarrow S \setminus \{s_{\max}, s'_{\min}, s_{\min}\}$\;
                    \uIf{$s' > s_{\min}$}{$M \leftarrow M \cup \{s'\}$\;}
                    \uElse{$L \leftarrow L \cup \{s'\} $\;} 
                }
                \uElse{$M \leftarrow M \cup \{e_1, e_2\}$\;}
                $E' \leftarrow E \setminus S$\;
            }
            $L \leftarrow L \cup \{s_{\min}\}$, $U \leftarrow U \cup \{s_{\max}\}$\;
            $ E'' \leftarrow \{ U, M, L \} $\;
            \Return{$E''$}
        }
    }
    \caption{{\bf IBWS} \label{alg:IBWS}}
\end{algorithm}
}

\subsection{Learning-to-Rank Model}
\label{ltr-model}
To predict the annotations from IBWS ranking, we train an automated scoring LTR model using data with annotated scores. Specifically, the model predicts an output $y \in [0, 1]$. We sample sentence pairs ($r_1$, $r_2$) where $r_1$ is annotated more positive than $r_2$ $(s_1 > s_2)$. A pairwise hinge loss with parameterized margin is used to train the model,
    \begin{equation}
    \max\{0, s_2 - s_1 + \alpha \cdot (f(r_1) - f(r_2))\}
    \end{equation}  
where $\alpha$ is the constant margin, $s_1, s_2$ is the annotated sentiment score and $f$ is the ranking model's score prediction function. The loss encourages the model to score $r_1$ higher than $r_2$. 

\paragraph{Pair Group Strategy} Considering that annotators may be more calibrated on a per-HIT or per-worker basis than on a global basis when using Amazon MTurk annotation~\citep{chen2020ranking}, we design pair grouping strategies, targeting to alleviate the disagreement between annotators and the inconsistency among tasks performed by the same annotator:

\begin{itemize}
    \item \textbf{Global basis} With n annotations, each one is paired with k randomly selected samples, maintaining a total of  $k \times n$ pairs. 
    % This strategy leads to a balanced spread of distances under the different protocols observed in the annotations.
    % means a fixed number of training examples, whether you have 4000 or 500 annotations.
    
    \item \textbf{Group by HIT} Samples are grouped by the \textit{HITId} to guarantee only pairs that are annotated in the same HIT by the same worker are used as training data. 
    % This is designed to mitigate both the inter-annotator and intra-annotator inconsistency.  
    \item \textbf{Group by worker} To reduce the impact of differences between annotators, samples are grouped by \textit{WorkerId} to guarantee only pairs annotated by the same worker are used as training data.    
\end{itemize}

% \subsection{Learning-to-Rank}
% \label{ltr-model}
% For experiment 4, we collected another 4k annotation (the same inputs used in the IBWS experiment) with 3-way redundancy using ``gold'' scalar protocol, slider interface, to explore the influence of the number of annotations and different pairwise approaches to the model's ranking ability. 

% To discuss the relationship between the number of annotations and ranking correlation, we randomly sample different n annotations repeatedly for ten iterations. The n is set among 4k, 2k, 1k, and 500. 

% There are $\frac{n(n-1)}{2}$ pairs to be considered globally, which is too expensive for training. We adopt a new random sampling strategy: given the number of annotations n, at iteration i, we generate a fixed number of input pairs for comparison of model performance when feeding different annotation numbers. Here the number is 36k (when n = 2k, each annotation is paired randomly with other 18 examples, when n = 4k, it's 10 pairs per annotation).
% For per-HIT / per worker basis, all the input pairs are included as training samples. 

\section{Experiments}
\label{Experiments}

% The following requirements qualify workers: HIT approval rate $>$ 98\%, location in the US, and number of HITs approved $>$ 100. To ensure that the worker population is consistent, we published all tasks at the same time on weekdays. 
% All tasks in Experiments 1 are published between 9:00 - 9:30 on Tuesday. For Experiment 2 and 3, we published 12 tasks consisting of 4 product categories in four weekdays. Each category contains 3 tasks, which are published one by one starting between 9:00 - 9:30. For each product category task, we blacklisted those workers who participated in previous tasks.

\subsection{Data}
We randomly select reviews from the Amazon product review dataset\footnote{https://nijianmo.github.io/amazon/index.html} \citep{Ni2019JustifyingRU}, ranging from four different product categories: \emph{Books, Electronics, Grocery-and-Gourmet-Food, and Home-and-Kitchen}. Each review covers information about the rating (1-5 stars), review text, product id, and reviewer id. 

% \begin{table}[t]
% \begin{center}
% \resizebox{0.4\textwidth}{!}{%
% \begin{tabular}{l|cc} 
% \toprule
% \textbf{Interface} & \textbf{IBWS} & \textbf{Scalar}\\
% \midrule
% \texttt{\#} annotations per item & 1 & 10\\
% \texttt{\#} items (4 types) & 1000x4 & 25x4\\
% \midrule 
% \texttt{\#} total annotations & 4000 & 1000 \\
% \bottomrule
% \end{tabular}
% }
% \caption{Annotation statistics. For each product type and each rating level, we randomly select 5 reviews for scalar interfaces, 200 reviews for IBWS annotation. To ensure a fair comparison, reviews selected for scalar annotation is included in IBWS reviews.}
% \label{tab:annotations}
% \end{center}
% \end{table}

\begin{figure}[t]
\vspace{-2mm}
\begin{center}
\includegraphics[width=1\linewidth]{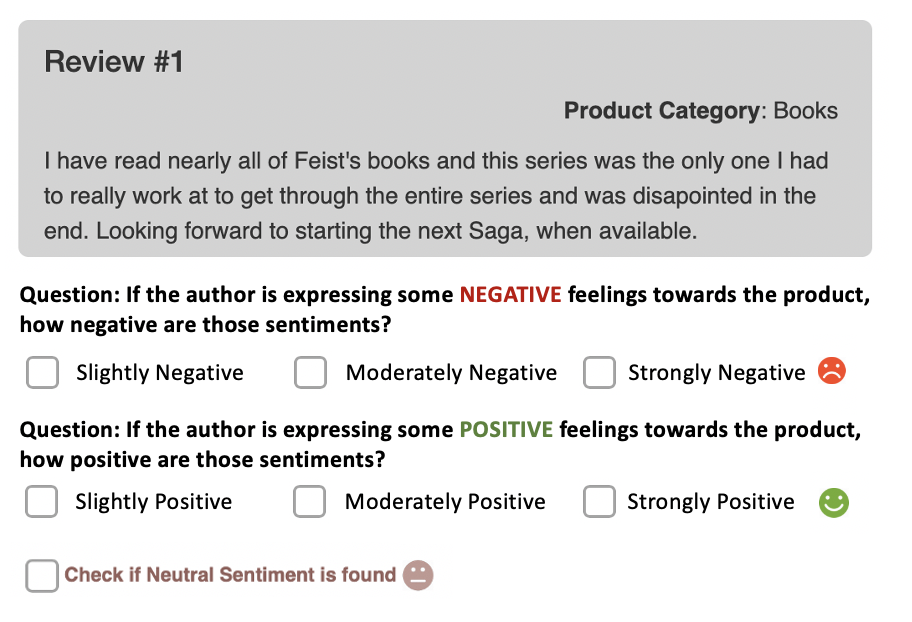}
\end{center}
\caption{Likert Style, dual-question Protocols.}
\label{fig:dual1}
\end{figure}

\subsection{Collecting Annotations}
We perform annotation collection with Amazon Mechanical Turk (AMT).\footnote{Full details of annotation task is in Supplementary Material.} 
% Table \ref{tab:annotations} presents the full annotation statistics. 

\paragraph{Direct Assessment} In an attempt to find the most robust and reliable scalar annotation protocols, we compare the collected annotations from \emph{7-way ordinal, slider} and \emph{VAS} protocols on 100 sampled product reviews. For each review, we collect 10 annotations, resulting in 1000 annotations for each protocol. Inspired by \citet{Yrizarry1998AmericanJapaneseDI}, we design \textbf{dual-category protocols} to examine if using separate scales for positive and negative sentiments improves annotation reliability. As shown in Figure \ref{fig:dual1}, these protocols allow annotators to select from either positive or negative sentiment categories (i.e., \emph{Neutral Sentiment} option is also available). In total, we collect annotations through six interfaces; each review will be presented either as a \emph{single question} or in a \emph{dual question} format.

% For single-question interface, we follow the design shown in Figure \ref{fig:single}. In terms of dual-question interfaces, we use the design shown Figure \ref{fig:dual1} as an ordinal scale. The same strategy is practiced in the slider and VAS scales, where we ask workers to annotate a value on either the positive bar (ranging from 50 to 100) or the negative bar (ranging from 0 to 50)\footnote{An example is shown in Appendix Figure \ref{fig:dual2}. To avoid a potential bias from consistently seeing the positive question above the negative one, the order of these two questions is randomized per webview.}.

% \citet{Yrizarry1998AmericanJapaneseDI} has demonstrated the validity of multiscalar rating in evaluating perceived emotion intensity. 

% Separating the positive and negative scale gives users more space to annotate and possibly mitigate this bias. This strategy is efficient given the different emotions described in the context. For example, given a sentence as follows:

% \begin{quote} \small
%     \emph{These sunglasses look good, but are uncomfortable to wear.}
% \end{quote}

% Both positive and negative sentiments are present as the author reviews two distinct attributes of the product. The annotator is unable to tell which feature the original reviewer considers more important. 

\paragraph{IBWS} We collect 4k annotations through two BWS interfaces with 3 iterations (include 100 reviews for direct assessment); only one worker is assigned to each task. The collected annotations are then ranked into 27 buckets and normalized to a [0, 1] scale (0 represents the most negative). 
% By comparing 100 reviews from the scalar annotations to the final IBWS-sorted list, we can evaluate which scalar interface aligns most closely with the IBWS ranking.

\subsection{Training the LTR Model} The \emph{best} scalar protocol determined from scalar annotation experiment is used to annotate training data: we collect another 4k annotations with 3-way redundancy to train an LTR model by fine-tuning the pre-trained RoBERTa \textsc{base} model.\footnote{Training details are in Supplementary Material.}
% to explore the influence of the number of annotations and different pairwise approaches on the model's ranking ability.

\paragraph{Metrics}
We evaluate the performance of RoBERTa-LTR models by computing the Spearman's rank correlation ($\rho$)~\citep{spearman04} between IBWS ranking and LTR predictions to test how closely the model's rankings align with the IBWS annotations. Intra-class correlation coefficient (ICC) \citep{Shrout1979IntraclassCU} is used to evaluate the reliability of annotators.

\subsection{Dialogue System Evaluation Experiments}

\begin{figure}[t]
\vspace{-2mm}
\centering
\includegraphics[width=1\linewidth]{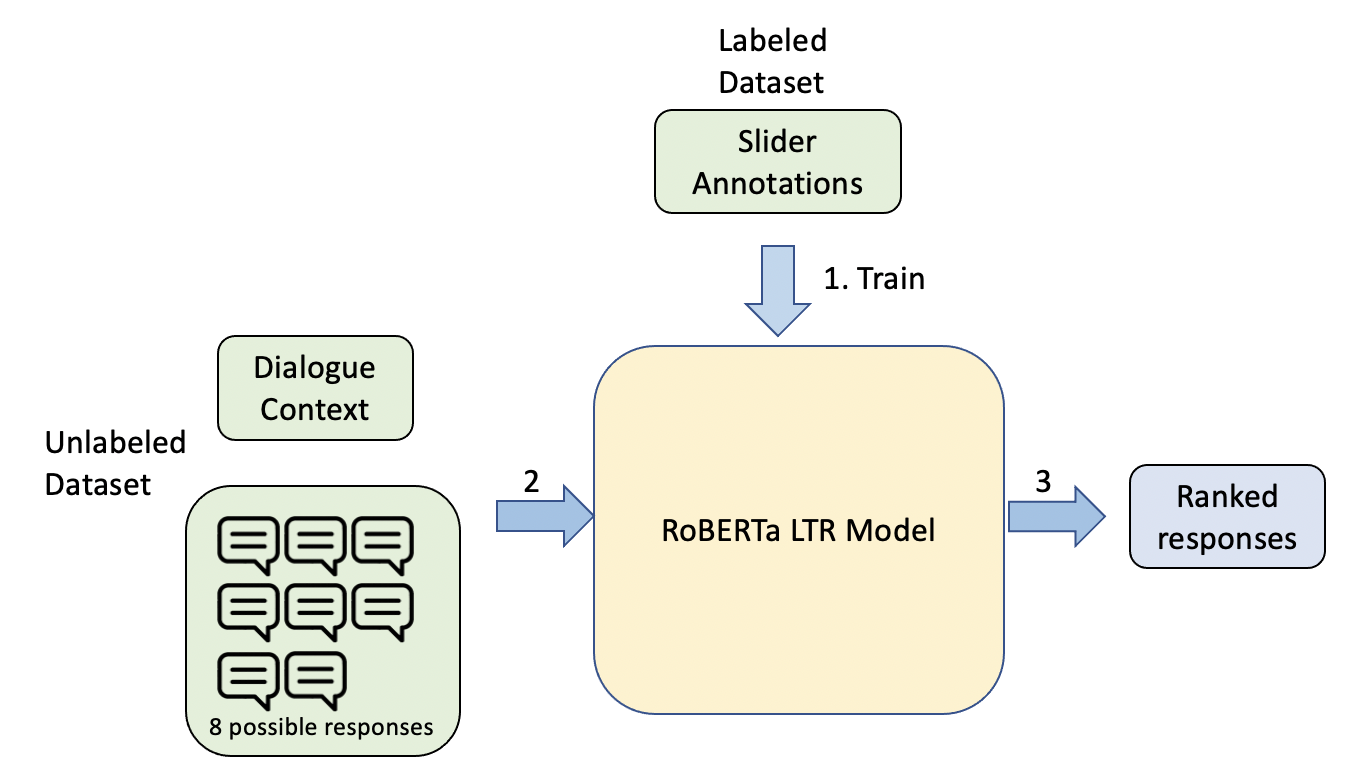}
\caption{LTR model on dialogue system outputs.}
\vspace{-4mm}
\label{fig:dialogue-model}
\end{figure} 

% \begin{figure*}[t]
% \centering
% \includegraphics[width=0.9\textwidth]
% {fig/analysis2.png}
% \caption{IBWS annotations correlate with ground truth labels (mean) in each bucket (round 3). Top: vertical-drag IBWS interface; Bottom: typical two column interfaces.}
% \label{fig:mean-ordinal}
% \end{figure*} 

\paragraph{Data} The dialogue data consists of 200 contexts and 40 responses for each context, for a total of 8000 context-response pairs. Each context has two conversational partners (A and B) speaking in turn, with A's sentence first, then B's response to A, and A's response to B (i.e. A-B-A). Each response is either a human-generated or a model-generated response from B to the last line of the conversation. For each context, 9 of the responses are written by humans, and 31 of the responses are generated by models. We use \textsc{CakeChat}\footnote{https://replika.ai/}; \textsc{DialoGPT (medium)}~\citep{zhang2020dialogpt}; \textsc{ConvAI2 (KV-MemNN)}~\citep{dinan2019second}; \textsc{Blender} (single turn); \textsc{Blender 2.7B}~\citep{roller2020recipes} with Person; \textsc{ParlAI} (Twitter 2); \textsc{ParlAI} (controllable)~\citep{see2019makes}; and \textsc{Plato-2}~\citep{bao2021plato2} (24 separate responses, from temperature $\{0.8, 0.9, 1.0\}$; top $k$ beam search size $\{10, 40\}$ or top $p$ beam search $\{0.8, 0.9\}$; and 2 responses per set of model parameters). 

\paragraph{Annotations}
Slider protocol is used to annotate the context-response pairs with the same setup. A subset of 2k context-response pairs was annotated with 3-way redundancy, while the rest (6k) were annotated without redundancy.

\paragraph{LTR Model}
The same model is used to train on the context-response pairs. The context-response pairs were spliced with RoBERTa's sentence separator token to form training and evaluation items. A total of 16 models are trained, as described below. Of the 200 contexts, 120 are set aside for training, 40 for the dev set, and 40 for the test set. Of the $120 \cdot 40 = 4800$ training context-response items, for each of the models, half of the items are chosen by one of the following data splits: the \textit{response} split, which has 60 contexts and 40 responses per context; the \textit{context} split, which has 120 contexts and 20 responses per context; the \textit{worker95} split, which contains a random sample of $2400$ items after filtering out annotations from the bottom 5\% of workers; or the \textit{worker80} split, which contains a random sample of $2400$ items after filtering out annotations from the bottom 20\% of workers. The ``bottom'' percentage of workers is determined as follows: for each worker, a correlation score can be computed for each context-response pair that was in the subset of pairs annotated with redundancy; in particular, the correlation between the worker's annotations and the mean of the other two worker's annotations for each redundant pair is computed. The workers are then sorted by their correlation scores, and the annotations of the workers in the bottom 5\% or 20\% are filtered. For each data split, 4 models are trained on the same pairwise hinge loss function, except instead of grouping by the \textit{HITId}, samples are grouped by the \textit{ContextId}.

\begin{figure}[t]
\centering
\includegraphics[width=1\linewidth]{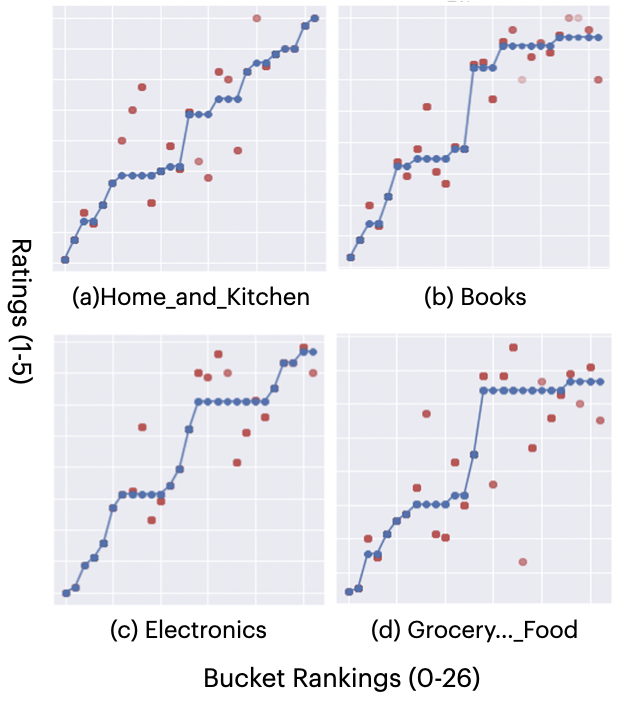}
\caption{Normalized IBWS annotations correlate with average ground truth labels.}
\vspace{-3mm}
\label{fig:mean-ordinal}
\end{figure} 

\begin{figure}[t]
\centering
\vspace{-2mm}
\includegraphics[width=1\linewidth]{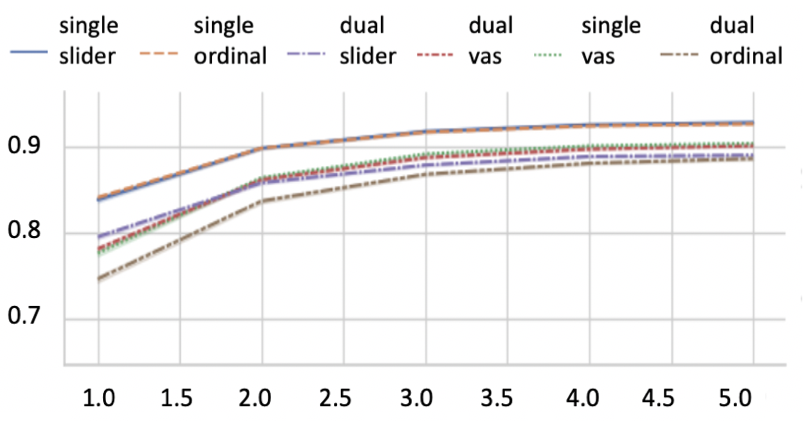}
\caption{Spearman's correlation of random split-half rankings. From top to bottom: single slider, single ordinal, dual slider, dual VAS, single VAS, dual ordinal.}
\label{fig:split-half}
\end{figure}

\begin{figure}[t]
\vspace{-3mm}
\includegraphics[width=1\linewidth]{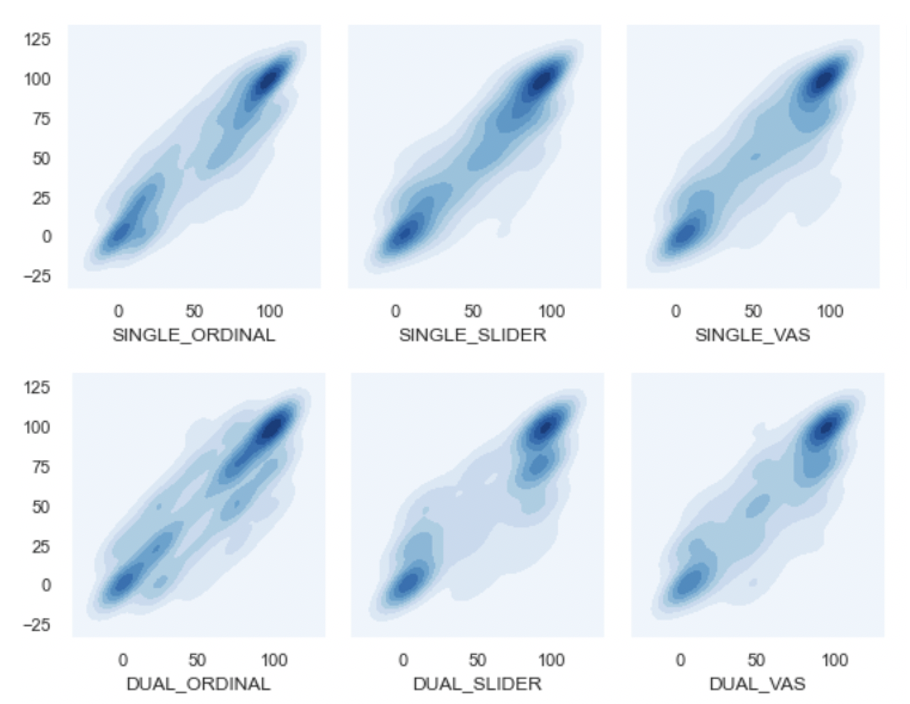}
\caption{Heatmaps of annotated rating score correlating with ground truth across the scalar interfaces.}
\vspace{-4mm}
\label{fig:heatmap}
\end{figure}

\begin{table}[t]
\centering
\resizebox{1\linewidth}{!}{%
\begin{tabular}{cccccc}
\toprule
 &  & ICC1 & ICC3 & ICC1k & ICC3k \\ \midrule
\multicolumn{1}{c}{Single} & \multicolumn{1}{c|}{Ordinal} & \hlc[violet!10]{\bf{0.74}} & 0.77 & \hlc[violet!10]{\bf{0.96}} & \hlc[violet!10]{\bf{0.97}} \\
\multicolumn{1}{c}{} & \multicolumn{1}{c|}{Slider} & \hlc[violet!10]{\bf{0.74}} & \hlc[violet!10]{\bf{0.78}} & \hlc[violet!10]{\bf{0.96}} & \hlc[violet!10]{\bf{0.97}} \\
\multicolumn{1}{c}{} & \multicolumn{1}{c|}{VAS} & 0.64 & 0.68 & 0.94 & 0.95 \\ \midrule
\multicolumn{1}{c}{Dual} & \multicolumn{1}{c|}{Ordinal} & 0.60 & 0.62 & 0.92 & 0.92 \\
\multicolumn{1}{c}{} & \multicolumn{1}{c|}{Slider} & 0.65 & 0.66 & 0.94 & 0.95 \\
\multicolumn{1}{c}{} & \multicolumn{1}{c|}{VAS} & 0.65 & 0.66 & 0.93 & 0.94\\ \bottomrule
\end{tabular}
}
\caption{ICC scores on annotations across all scalar protocols.}
\vspace{-6mm}
\label{tab:icc-scalar}
\end{table}

\begin{table}[t]
\begin{center}
\resizebox{1\linewidth}{!}{%
\begin{tabular}{cccccc}
\toprule
% & \multicolumn{6}{c}{$\rho$} \\ \cline{2-7} 
 \multicolumn{3}{c|}{Single} & \multicolumn{3}{c}{Dual} \\ 
 Ordinal & Slider & \multicolumn{1}{c|}{VAS} & Ordinal & Slider & VAS \\ \midrule
% \multicolumn{1}{c|}{\textbf{median}} & \multicolumn{1}{c|}{0.872} & \multicolumn{1}{c|}{0.875} & \multicolumn{1}{c|}{0.872} & \multicolumn{1}{c|}{0.845} & \multicolumn{1}{c|}{0.880} & 0.881 \\ \hline
\multicolumn{1}{c}{\hlc[violet!10]{\bf{0.881}}} & \multicolumn{1}{c}{\hlc[violet!10]{\bf{0.881}}} & \multicolumn{1}{c|}{0.877} & \multicolumn{1}{c}{0.828} & 
\multicolumn{1}{c}{0.872} & 0.879 \\ \bottomrule
\end{tabular}
}
\end{center}
\caption{Spearman correlation ($\rho$) between scalar annotations and true labels.}
\vspace{-3mm}
\label{tab:true-label-corr}
\end{table}

\section{Results and Analysis}
\label{Results}

\subsection{The Effectiveness of IBWS}
To evaluate the reliability of IBWS and confirm its inheritance of BWS's robustness, we compute the Spearman's correlation between the rankings generated from IBWS and the average true ordinal labels within each bucket, as depicted in Figure \ref{fig:mean-ordinal}. The observed consistent, monotonically increasing trend across all product types confirms that IBWS effectively ranks the reviews as intended.

However, several factors contribute to why the plots are not perfectly sorted: 1) poor-quality responses; 2) annotators might focus on different aspects than the ground-truth ratings (e.g., prioritizing certain attributes that differ from those emphasized by other reviewers); and 3) the buckets may not align in a strictly linear fashion with the ground truth ratings.

By comparing the results from the standard two-column BWS interface and the vertical-drag interface (See Supplementary), we find that annotators performed better with the standard two-column setup. Although both interfaces produce a monotonic relationship, the annotations from the standard two-column interface show less variance relative to the true ordinal ratings in each bucket. Additionally, the vertical-drag interface results in more outliers being misclassified.

% Figure \ref{fig:bws-worktime} compares the average annotation time per assignment used in the standard BWS interface and the vertical-drag interface on Amazon review product sentiment annotations, broken down by product categories. The mean time for standard BWS is 145 seconds, while the mean time for the vertical-drag BWS is 362.5 seconds. We published the tasks in the order of Books, Electronics, Grocery \& Food, and Home \& Kitchen. Both interfaces' average time gradually increase with this order as well. We suspect two reasons that may have caused this: 1) After we added workers to our blacklist after each product type task, there are less efficient AMT workers. 2) The annotations were performed close before the Christmas holiday; when approaching a holiday, the workers may not be as efficient as on regular workdays.

% \begin{figure}[h]
% \includegraphics[width=1\linewidth]{fig/bws-worktime.png}
% \caption{Average work time in seconds of BWS interfaces.}
% \label{fig:bws-worktime}
% \end{figure} 

\subsection{Rating Scale Performance}
\paragraph{Reliability and Stability}
To examine how consistently we get similar rankings from every direct assessment protocol, we employ the random \emph{split-half}~\citep{kiritchenko-mohammad-2017-best} with Spearman's correlation score. Specifically, for each review, we randomly sample two annotations out of ten to form lists A and B respectively. Ties in the resulting rankings are broken by adding a small amount of random noise, and Spearman's correlation is computed between A and B. As illustrated in Figure \ref{fig:split-half}, single slider and single ordinal protocols yield the highest consistency.

% To get a sense of each annotation protocol's reliability when it does not depend on having a gold standard to compare, we employ the random \emph{split-half} method described by \citet{kiritchenko-mohammad-2017-best} to examine how consistently we get similar rankings from the various protocols. Specifically, for each reviews, we randomly select two annotations out of ten to form list A and B respectively.  Ties in the resulting rankings are broken by adding a small amount of random noise, and Spearman's correlation is computed between A and B. As illustrated in Figure \ref{fig:split-half}, single slider and single ordinal protocols yield the highest consistency.

Table \ref{tab:icc-scalar} presents the ICC across various scalar annotation protocols.  The results align with the findings from Figure \ref{fig:split-half}; the single-category ordinal and slider interfaces perform more reliably and efficiently than the others.

\paragraph{Effectiveness} 

Table \ref{tab:true-label-corr} compares the correlation of each scalar annotations (i.e., the mean of 10 redundant annotations) with the \textbf{ground truth} values from the original Amazon review dataset. All three single-category interfaces outperform the dual-category ones, with the single slider and ordinal scales showing better correlation than the VAS scale. Additionally, as shown in Figure \ref{fig:heatmap}, the single slider annotations are most concentrated along the diagonal, indicating the strongest alignment with the ground-truth labels.

\begin{figure}[t]
\centering
\includegraphics[width=1\linewidth]{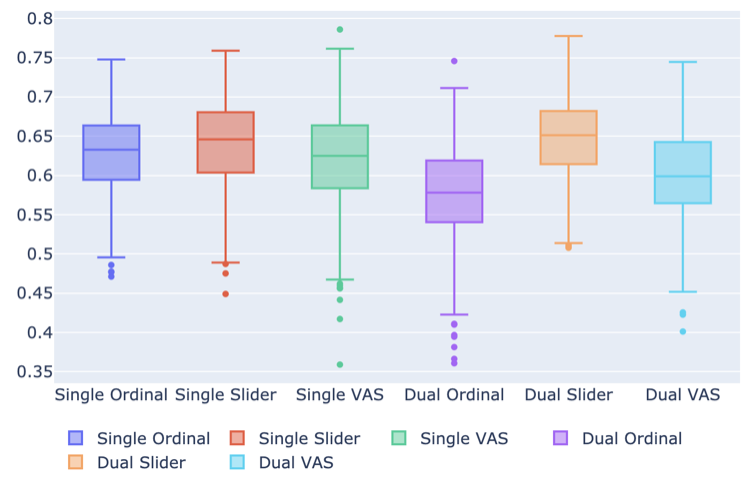}
\caption{Spearman's correlation ($\rho$) across scalar interfaces with IBWS annotations at zero redundancy (AR = 1).}
\vspace{-2mm}
\label{fig:no-redundancy}
\end{figure} 

\begin{figure}[t]
\centering
\includegraphics[width=1\linewidth]{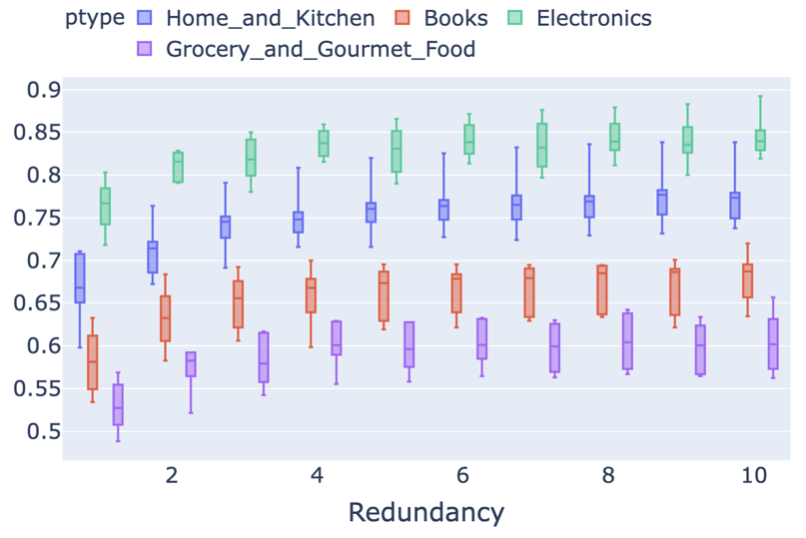}
\caption{Median Spearman's correlation ($\rho$) between IBWS and scalar annotations across protocols and product types.}
\vspace{-3mm}
\label{fig:redundancy_product}
\end{figure}

\begin{figure}[t]
\includegraphics[width=1\linewidth]{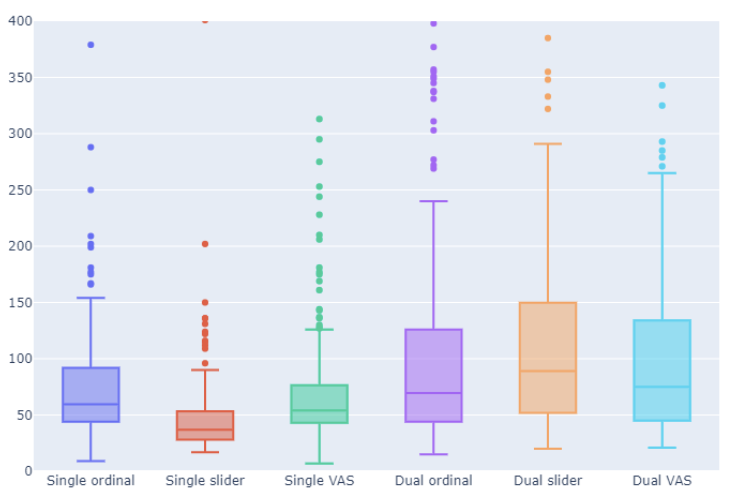}
\caption{Work time of scalar interfaces in seconds.}
\vspace{-3mm}
\label{fig:scalar-worktime}
\end{figure} 

% \paragraph{Which protocol gets the most reliable and stable annotations, considering ties in the computation of correlation?}

% \paragraph{IBWS Correlations}
Figure \ref{fig:no-redundancy} illustrates the correlation between \textbf{IBWS-ranked annotations} and scalar annotations, where a single annotation is randomly selected from the 10 redundant annotations for each review. At zero redundancy (i.e., when only one annotator's input is considered), the slider interfaces show noticeably higher correlations compared to the other two types. We also find that when redundancy increases, the correlation gradually increases across all scalar protocols, and all product types, as shown in Figure \ref{fig:redundancy_product}.

% \paragraph{Annotation Redundancy} To explore the impact of annotation redundancy (AR) on the results, we randomly draw $n$ annotations for each HIT at each redundancy level between 1 and 10. We then compute the median rating scores by aggregating the selected annotations and correlate them with the normalized IBWS ranking scores. This process is repeated 100 times to ensure balanced results. The average correlations per redundancy level and protocol are shown in Figure \ref{fig:redundancy_corr}. When $AR < 3$, slider protocols show the highest correlation with IBWS rankings. With redundancy increases, VAS protocols gradually perform better. Overall, sliders consistently align well with both IBWS and the ground truth.

% \paragraph{Product Type} Figure \ref{fig:redundancy_product} shows how different product types impact the correlation at each redundancy level. Table \ref{tab:icc-ptype} presents the ICC across different product types. Based on these observations, we conclude that the differences between annotation protocols are smaller than the differences between product categories in the scenarios we examined.

% \begin{figure}[ht]
% \centering
% \includegraphics[width=0.5\textwidth]{fig/corr_ptype.png}
% \caption{Median Spearman’s correlation between standard BWS and scalar interfaces across all product types.}
% \label{fig:redundancy_product}
% \end{figure}

\paragraph{Efficiency} Figure \ref{fig:scalar-worktime} plots the annotation time taken by workers to rate 5 reviews across all scalar interfaces. Overall, the single slider method was most efficient for the sentiment annotation task. The dual interfaces all took longer than their single counterparts.

\begin{table*}[t]
\centering
\resizebox{0.8\linewidth}{!}{%
\begin{tabular}{lcccccccc}
\toprule
 Model / Data Split & \multicolumn{2}{c}{response} & \multicolumn{2}{c}{context} & \multicolumn{2}{c}{worker95} & \multicolumn{2}{c}{worker80} \\
            & dev   & test  & dev   & test  & dev   & test  & dev   & test  \\  \midrule
pointwise   & \hlc[violet!10]{\bf{35.04}} & \hlc[violet!10]{\bf{32.17}} & \hlc[violet!10]{\bf{35.91}} & \hlc[violet!10]{\bf{31.68}} & 32.84 & 30.41 & \hlc[violet!10]{\bf{35.59}} & 31.93 \\
global      & 27.92 & 26.88 & 30.20 & 28.33 & 31.63 & 26.81 & 31.97 & 31.26 \\
per-worker  & 29.46 & 26.87 & 33.52 & 29.64 & 34.09 & 29.36 & 32.12 & 29.79 \\
per-context & 33.16 & 30.92 & 35.59 & 30.93 & \hlc[violet!10]{\bf{35.50}} & \hlc[violet!10]{\bf{31.97}} & 35.53 & \hlc[violet!10]{\bf{31.99}} \\
\bottomrule
\end{tabular}
}
\caption{Spearman's correlation ($\rho$) of RoBERTa-LTR models trained and evaluated on dialogue data.}
\vspace{-4mm}
\label{tab:ltr-dialogue}
\end{table*}

\begin{table}[h]
\resizebox{1\linewidth}{!}{%
\centering
\begin{tabular}{lllll}
\toprule
% &  & \multicolumn{3}{l}{$\rho \times 100$} \\ \cline{2-5} 
Model / Training size & 500 & 1000 & 2000 & 4000 \\ \midrule
global & \hlc[violet!10]{\textbf{66.29}} & \hlc[violet!10]{\textbf{69.94}} & \hlc[violet!10]{\textbf{71.86}} & \hlc[violet!10]{\textbf{72.56}} \\
per-HIT & 60.43 & 65.57 & 71.58 & 72.52 \\
per-worker & 59.45 & 64.39 & 70.92 & 72.18  \\ \bottomrule
\end{tabular}
}
\caption{Spearman's correlation ($\rho$) of RoBERTa-LTR model predictions, evaluated on IBWS sentiment annotations.}
\vspace{-4mm}
\label{tab:ltr-corr}
\end{table}

\subsection{LTR Model Performance}

Table \ref{tab:ltr-corr} presents the performance of the RoBERTa-LTR model trained on different sizes of sentiment annotations collected from the slider protocol (500, 1k, 2k, and 4k) and tested on the 4k IBWS annotations. We observe that as the number of training annotations increases, the prediction accuracy improves, and the performance gap between the three pairwise strategies narrows. When the training dataset is small, the global pairwise strategy significantly outperforms the other two settings, likely due to the difference in the number of training pairs. However, this approach requires six times more training time. Once the model is trained on more than 2,000 annotations, all approaches—global, per-context, and per-worker—achieve a correlation accuracy above 0.7, indicating the model is well-trained to predict rankings.

\paragraph{Performance on dialogue dataset}
Table \ref{tab:ltr-dialogue} shows that the RoBERTa-LTR model achieves a Spearman's correlation of \~0.3 on dialogue annotations, which is significantly lower than on sentiment data. However, the inter-annotator correlation on the redundantly-annotated subset is also 0.3, implying that the models are approaching human performance.

% Among the dialogue models, per-worker models performed similarly to global models, while pointwise models matched per-context models, with per-context outperforming pointwise. Data splits had minimal impact on performance, given the random noise from training input order.
With regards to the performance between the dialogue models, across the various data splits, the per-worker models tend to perform on par with the global models, while the pointwise models perform on par with the per-context models, with the latter pair outperforming the former. The data split is not found to have a significant effect on model performance, especially with the random noise in performance from the randomized order of the training inputs.

\section{Related Work}
\label{Background and Related work}

% In this section, we introduce related work in terms of annotation reliability and automated scoring models.
% Within natural language processing sentiment intensity and language generation system ranking are tasks with a large amount of annotation effort that focuses on rating items on a scale \cite{novikova-etal-2018-rankme,kiritchenko-mohammad-2017-best, sakaguchi-van-durme-2018-efficient}. 

\paragraph{Annotation Reliability}
% Annotation reliability is critical in training and evaluating machine learning models.
% The most commonly used direct assessment methods are the $n$-way ordinal scale \citep{likert_1987} and slider scale, as shown in Fig \ref{fig:single}. 
% The slider scale requires annotators to drag a slider handle to indicate their assessment, while a similar tool, the Visual Analog Scale (VAS) \citep{hayes1921experimental}, simply asks users to click on a horizontal bar. 
\citet{Amidei2019TheUO} points out a lack of robustness studies on the use of ordinal scales in natural language generation evaluations. Previous research comparing different direct assessment protocols, such as ordinal, slider, VAS, and swipe (a mobile-friendly variant of the slider), finds minimal statistical differences in data reliability and completion times in web and self-report surveys \citep{Fryer2020TheFO, web}. The reliability and robustness of scalar annotations remain uncertain. On the other hand, BWS, which relies on relative comparisons, has been shown to produce more accurate and reliable sentiment intensity annotations compared to direct assessment methods \citep{kiritchenko-mohammad-2017-best}. Additionally, BWS is effective in fields such as psychology \citep{e1c6ab23f39d4335b1f68a87ac7105f8}, NLP data annotation \citep{van-miltenburg-etal-2023-reproducible}, and has even been suggested as a replacement for ordinal scales in healthcare experiments \citep{FLYNN2007171}.

Several strategies have been proposed to improve annotation reliability and consistency. For example, Efficient Annotation of Scalar Labels (EASL) combines direct assessment with online pairwise ranking aggregation \citep{efficient-online-scalar-annotation-with-bounded-support}. Rank-Based Magnitude Estimation (RankME) integrates continuous scales with relative assessments to enhance the reliability and consistency of human ratings \citep{Novikova_2018}. \citet{santhanam-shaikh-2019-towards}  compares four experimental designs—Likert scale, RankME, BWS, and Biased Magnitude Estimation (BME)—in evaluating dialogue systems based on readability and coherence. 
% They found that BWS was less reliable than continuous estimation methods. 
% Further research by \citet{Santhanam2020StudyingTE} examined the impact of anchoring bias on dialogue evaluation, emphasizing the need for robust annotation protocols.
%EASL 
%RANKME
%Samira - dialog 
% \citet{e1c6ab23f39d4335b1f68a87ac7105f8} shows a case where BWS is employed in psychological research for first impression measurement. The study begins by showing annotators four pictures of facial expressions instead of text samples which are more commonly used with BWS. 

% \citet{louviere_flynn_marley_2015} also discusses that the BWS task is well-grounded and straightforward, but in the worst case, requires more time than categorical rating scales. Furthermore, \citet{glenn-etal-2022-viability} indicates that BWS approximately takes 4.5 times longer than categorical annotation.

% \paragraph{Ranking Annotations}
% \citet{Orme2009MaxDiffA} explores counting analysis to rank a complete set of BWS-annotated items, where each item's score is calculated by subtracting the fraction of times it is selected as the worst from the fraction of times as the best. \citet{mohankumar2022activeevaluationefficientnlg} introduces active evaluations to identify the top-ranked system efficiently.

\paragraph{Automated Scoring Models}
\citet{Orme2009MaxDiffA} explores counting analysis to rank a complete set of BWS-annotated items. \citet{mohankumar2022activeevaluationefficientnlg} introduces active evaluations to identify the top-ranked system efficiently.
Learning-to-rank has been extensively used in various NLP tasks, such as ranking candidate translations for a given sentence, determining document relevance to a query, and ranking sentences by sentiment intensity, as in our study \citep{li2023learning, yan2023learning, Frydenlund_Singh_Rudzicz_2022, luo2023modelbased}. \citet{Ekbal_2011} introduce AugSBERT that improves pairwise sentence scoring tasks on crowdsourced data annotated via the BWS interface. Previous research has enhanced state-of-the-art performance by incorporating pre-trained models like BERT \citep{devlin2019bert} and RoBERTa \citep{liu2019roberta} into LTR frameworks, with findings showing that RoBERTa slightly outperforms BERT in ranking tasks \citep{han2020learningtorank}. \citet{tang-etal-2022-investigating} integrates the Likert scale into BWS, where annotators assess the distance between the best and worst options using a 3-point ordinal scale to rank items for summarization factual consistency.
\section{Conclusion}
Best Worst Scaling (BWS) is a respected annotation procedure on small datasets. We introduced Iterated BWS as a robust method for crowdsourced annotation of larger collections. While robust, IBWS requires repeated consideration of each element in the collection: $k$ iterations translates to $k$ times the cost. We illustrated that a direct scalar assessment of each element using a slider protocol allows for significantly more efficient annotation, while giving similar results to IBWS.
These annotations support training automated pairwise ranking models: in both sentiment analysis and dialogue tasks, the LTR models effectively predict rankings on par with human annotations.
To our knowledge, this study is the first to directly consider the widely regarded BWS protocol in the context of large datasets and with an eye to practical considerations of annotation costs. Our results support the conclusion that researchers can comfortably rely on a direct scalar assessment protocol as a more efficient and similarly robust approach. 

%Such protocols lead to similar rankings of elements, and can directly support automated scoring methods.

%Despite BWS requiring more evaluations than direct assessments, slider interfaces efficiently align with IBWS rankings using only O(n) interactions. Our analysis shows that single-category ordinal and slider interfaces are stable. We also explore an automated pairwise ranking model, RoBERTa LTR, with three strategies, finding that increased annotation volume improves performance.  In both sentiment analysis and dialogue tasks, the LTR models effectively predict rankings on par with human annotations.
% The per-HIT strategy, in particular, delivers superior results on dialogue datasets, providing a new approach for enhancing annotator consensus in crowdsourcing.
% In this paper, we introduce IBWS as a reliable ranking method for crowdsourcing data. We illustrate that despite BWS requiring more evaluations than direct assessments, slider interfaces are efficient and align well with IBWS rankings with only O(n) interactions needed. Our findings highlight the stability of single-category ordinal and slider interfaces through a split-half ranking analysis. Furthermore, we explore an automated pairwise ranking model, RoBERTa LTR, utilizing three distinct strategies. Our findings emphasize that increasing annotation volume enhances model performance. Notably, the per-HIT pairwise strategy achieves superior results on a dialogue dataset, offering a novel approach to improving annotator consensus in crowdsourcing data collection.

% \section{Acknowledgments}

\bibliography{aaai25}

\end{document}